\pdfoutput=1
\PassOptionsToPackage{table}{xcolor}
\documentclass[11pt]{article}
\usepackage[final]{acl}

\usepackage{times}
\usepackage{amssymb}
\usepackage{latexsym}
\usepackage{diagbox}
\usepackage{graphicx}
\usepackage{float}
\usepackage{hyperref}

\usepackage[T1]{fontenc}

\usepackage[utf8]{inputenc}
\usepackage{booktabs}
\usepackage{multirow}
\usepackage{lipsum}

\usepackage{multirow}
\usepackage{booktabs}
\usepackage{graphicx}
\usepackage[table]{xcolor}

\newcommand{\diagcell}[1]{\cellcolor{magenta!20}#1}

\usepackage{microtype}

\usepackage{inconsolata}

\usepackage{graphicx}

\title{Dialectal Coverage And Generalization in Arabic Speech Recognition}

\author{Amirbek Djanibekov$^{1}$\thanks{\phantom-These authors contributed equally to this work.}, Hawau Olamide Toyin$^{1}$\footnotemark[1] \\ {\bf Raghad Alshalan$^{2}$}, {\bf Abdullah Alitr$^{2}$}, {\bf Hanan Aldarmaki$^{1}$} \\ \\
$^{1}$Mohamed bin Zayed University of Artificial Intelligence, Abu Dhabi, UAE \\
$^{2}$STC, Riyadh, Saudi Arabia \\
\texttt{\fontsize{10}{12}\selectfont \{amirbek.djanibekov,hawau.toyin,hanan.aldarmaki\}@mbzuai.ac.ae}\\
\texttt{\fontsize{10}{12}\selectfont \{rsalshalan,aalatir\}@stc.com.sa}\\
}

\begin{document}
\maketitle
\begin{abstract}
    Developing robust automatic speech recognition (ASR) systems for Arabic 
    requires effective strategies to manage its diversity. Existing ASR systems mainly cover the modern standard Arabic (MSA) variety and few high-resource dialects, but fall short in coverage and generalization across the multitude of spoken variants. Code-switching with English and French is also common in different regions of the Arab world, which challenges the performance of monolingual Arabic models. In this work, we introduce a suite of ASR models optimized to effectively recognize multiple variants of spoken Arabic, including MSA, various dialects, and code-switching. We provide open-source pre-trained models that cover data from 17 Arabic-speaking countries, and fine-tuned MSA and dialectal ASR models that include at least 11 variants, as well as multi-lingual ASR models covering embedded languages in code-switched utterances. We evaluate ASR performance across these spoken varieties and demonstrate both coverage and performance gains compared to prior models. 

\end{abstract}

\section{Introduction}
\begin{figure*}
    \centering
    \includegraphics[width=1.6\columnwidth]{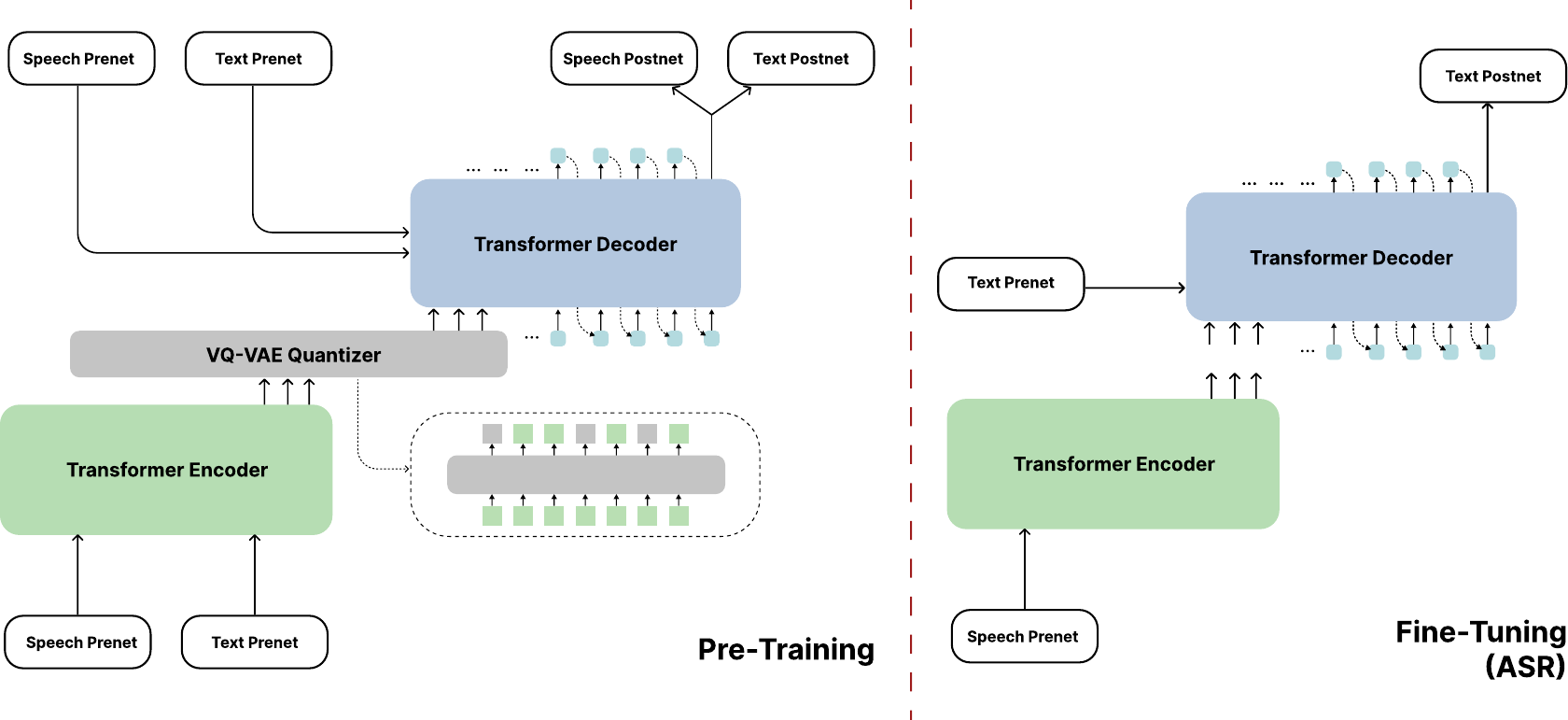}
    \caption{The architecture of SpeechT5/ArTST, which contains an encoder-decoder module and six modal specific pre/post-nets. During self-supervised pre-training (left), quantized tokens are shared across speech and text modalities. Hidden states and latent units are mixed up and used as the inputs of the cross-attention module in the decoder. The fine-tuning stage for ASR is shown on the right. Refer to \citet{ao-etal-2022-speecht5} for more details.}
    \label{fig:model_architecture}
\end{figure*}

The advent of large self-supervised acoustic models has transformed speech technology, 
enabling transfer learning and improving performance for both high-resource and low-resource languages. Prominent examples of such models include various versions of wav2vec~\cite{schneider2019wav2vec, baevski2020wav2vec}, HuBERT~\cite{hsu2021hubert}, and SpeechT5~\cite{ao-etal-2022-speecht5}, which have predominantly been trained on English datasets. 
Their multi-lingual variants, e.g. XLS-R~\cite{babu2022xls} with 53 and 128 languages, in addition to models that include both self-supervised and supervised pre-training, such as Whisper~\cite{radford2023robust} with approximately hundred supported languages, MMS \cite{pratap2024scaling} with thousands of languages, and UniSpeech~\cite{Wang2021UniSpeech}, illustrate the potential for cross-lingual transfer learning for more inclusive ASR. Yet, while these models indeed show great potential for transfer learning to new languages, even those unseen in training, they remain suboptimal for some target languages. 
A case in point is the Arabic Text and Speech Transformer (ArTST), a model pre-trained exclusively on Arabic, which has demonstrated superior performance for Modern Standard Arabic (MSA), surpassing larger multilingual models like Whisper and MMS in benchmark tests, in addition to establishing a new state-of-the-art performance compared to previous efforts for Arabic ASR. This highlights the advantage of monolingual pre-training when large amounts of unlabeled data for the target language are available.
While the model showed some potential for dialectal coverage, it was trained and validated exclusively on MSA data, which questions its applicability for spoken dialectal variants of Arabic. Evaluations on code-switched data showed poor performance of ArTST compared to multilingual models \cite{kadaoui-etal-2024-polywer}, demonstrating the delicate trade-off between monolingual and multilingual optimization. Arabic is a pluricentric language \cite{SCHUPPLER2024103007},  diverse in regional variations, and models trained on MSA frequently struggle to adapt to these variations. 
This limitation is particularly acute given that many Arabic dialects are underrepresented and considered low-resource in speech technology research. Consequently, there is a  need for optimized ASR systems that embrace, rather than overlook, the linguistic diversity of the Arabic-speaking world.

In light of these challenges, we conduct various investigations aimed at understanding and enhancing the dialectal diversity and performance of Arabic ASR systems. We focus on four inquiries aimed at optimizing potential strategies for integrating dialectal variation into ASR systems. 
First, we measure the impact of incorporating a broad collection of Arabic dialects during the model's pre-training phase. We hypothesize that a wider dialectal foundation could improve the model's performance across various dialects in the fine-tuning stage. 
Second, we quantify the comparative effectiveness of dialect-specific fine-tuning versus a more holistic, multi-dialectal fine-tuning strategy. 
The third question examines the model's capacity for zero-shot transfer to dialects not explicitly included in fine-tuning. Finally, we evaluate the model on code-switched utterances, and examine the effect of multilingual pre-training and fine-tuning on both monolingual and code-switched datasets. 
Our key findings from experiments spanning over 17 variants of spoken Arabic are: (1) pre-training with more data and wider dialectal coverage improves performance across most dialectal variants, including MSA, (2) multi-dialectal fine-tuning improves performance for low-resource dialects, but may not be optimal for high-resource dialects, (3) multi-dialectal pre-training and fine-tuning has higher potential for zero-shot transfer to unseen dialects, and (4) multi-lingual pre-training and fine-tuning greatly boosts performance on code-switching, while negatively impacting  monolingual performance due to language interference. Our pretraining checkpoints and joint models were trained exclusively on open-source data and are released as open-source, open-weights models. All scripts required to reproduce our results including model training, evaluation, and checkpoints, are publicly available~\footnote{\url{https://github.com/mbzuai-nlp/ArTST}}.

\section{Related Work}
Arabic speech recognition research has a long history, but the majority of this research has focused on Modern Standard Arabic (MSA), the formal variant predominant in news broadcasts and official communications. A review article covering peer-reviewed publications between 2011 and 2021 estimates that 89\% of papers on Arabic ASR cover MSA, and only a quarter cover some dialectal variant of Arabic \cite{dhouib2022arabic}. Recent research in end-to-end ASR for Arabic, as presented in~\citet{hussein2022arabic}, demonstrates the potential of contemporary deep learning techniques for decoding spoken Arabic, but it was also limited to MSA. Large-scale multilingual ASR models, such as Whisper~\cite{radford2023robust} and MMS~\cite{pratap2024scaling}, cover many languages within their scope, including Arabic. They utilize language embeddings or adapters to enhance language coverage and performance within the same model, but their performance across languages vary considerably. 
For instance, while Whisper demonstrates strong zero-shot capabilities on MSA, its zero-shot accuracy drops substantially on dialects, and additional finetuning on dialectal data is needed to improve performance~\cite{waheed-etal-2023-voxarabica}. Recent work has shown that smaller, Arabic-specific student models distilled from large models like Whisper can achieve comparable or even superior performance, especially on dialectal data, with good generalization to unseen dialects \cite{waheed2024distill}. Specialized Arabic models like ArTST~\cite{toyin-etal-2023-artst}, primarily pre-trained on MSA, have shown competitive results on MSA tasks and even outperformed larger multilingual models on some mixed-dialect datasets like QASR~\cite{mubarak-etal-2021-qasr}. However, due to its monolingual pretraining, the model was shown to perform poorly in code-switched Arabic-English speech~\cite{al-ali-aldarmaki-2024-mixat}. 
This illustrates that strong MSA performance is not a reliable predictor for dialectal or code-switching capabilities, with a substantial gap persisting between SOTA MSA and dialectal performance.
The development of diverse datasets such as QASR (multi-dialectal broadcast news) \cite{mubarak-etal-2021-qasr}, SADA (Saudi Arabic) \cite{10446243}, ArZen (Egyptian-English code-switching) \cite{al2024arzen}, and  Mixat (Emirati-English code-switching) \cite{al-ali-aldarmaki-2024-mixat}, and other public datasets covering various dialects and code-switched instances presents an opportunity for improving the generalization of ASR systems to diverse spoken varieties.

\section{ Methodology}

Based on prior work, we start with the premise that monolingual  training is more suitable for maximizing performance in Arabic ASR. However, the current Arabic SOTA models have limited coverage of spoken varieties and struggle with code-switching due to their monolingual training.   Our objective is to maximize performance while also widening the coverage to include MSA, regional dialects, and instances of code-switching. 
To that end, we start with an Arabic-centric ASR model, ArTST \cite{toyin-etal-2023-artst}, as the foundation for our investigation. Figure~\ref{fig:model_architecture} illustrates the high-level architecture of ArTST for self-supervised pre-training and fine-tuning. This model is based on the SpeechT5 approach \cite{ao-etal-2022-speecht5}, and supports multi-modal fine-tuning. 
The first version of the model was pre-trained on the MGB2~\cite{ali2016mgb} dataset, which consists of newswire data, mainly in MSA, with a small subset of dialectal variants. 
In this work, we attempt to understand the factors that enable both high performance and wide coverage; we explore the following questions:

\begin{enumerate}
\setlength{\parskip}{0pt}
    \item Is \textbf{pretraining} on dialectal data beneficial for improving down-stream dialectal performance, and would it negatively impact MSA performance?
    \item Is it better to \textbf{finetune} ASR models jointly on multiple dialects or fine-tune on a specific target dialect?
    \item Can we achieve reasonable \textbf{zero-shot} performance on unseen dialects?
     \item Can we optimize performance in \textbf{code-switched} utterances using multilingual pretraining?
    \item What is the effect of \textbf{multilingual} pretraining and finetuning on monolingual Arabic performance? (i.e. language interference).
    
\end{enumerate}

The remaining sections detail our experimental settings and findings of these questions.

\subsection{Terminology}
For the rest of the paper, we will refer to Arabic variants using abbreviations. The categories below are based on regions and countries, and \textit{do not reflect any official classification} of dialectal families: 

    \paragraph{\textbf{\textcolor{purple}{MSA:}}} Modern Standard Arabic. This is a common official variant of Arabic used in news, books, and education.  \textbf{\textcolor{purple}{CA}}: Classical Arabic. This is an old variant of Arabic found on religious texts and old books. It resembles MSA, but also contains outdated lexical items and structures. \paragraph{\textbf{\textcolor{purple}{GLF:}}} A broad category of dialects spoken in the Arabian Peninsula, in particular the Gulf region, which, in our data sources, include \textbf{SAU}: Saudi, \textbf{KUW}: Kuwait, \textbf{UAE}, \textbf{OMA}: Oman, \textbf{QAT}: Qatar, \textbf{IRA}: Iraq, and \textbf{YEM}: Yemen. 
    \paragraph{\textbf{\textcolor{purple}{LEV:}}} Levantine dialects,  which, in our data sources, include \textbf{SYR}: Syria, \textbf{JOR}: Jordan, \textbf{LEB}: Lebanon, and \textbf{PAL}: Palestine.  \paragraph{\textbf{\textcolor{purple}{NOR:}}} North African dialects, including \textbf{EGY}: Egypt,  \textbf{TUN}: Tunisia, \textbf{MOR}: Morocco, \textbf{ALG}: Algeria, \textbf{MAU}: Mauritania, and \textbf{SUD}: Sudanese.

\subsection{Pre-Training Data \& Settings}\label{sec:pre-train-data}

To examine the effect of pre-training data coverage on downstream performance, we pre-trained ArTST from scratch\footnote{We used the scripts and configurations provided in \url{github.com/mbzuai-nlp/ArTST}} on both MSA and dialectal data. We sourced our data from various datasets, including: MGB2~\cite{ali2016mgb}, QASR~\cite{mubarak-etal-2021-qasr} MGB3~\cite{ali2017speech}, MGB5~\cite{ali2019mgb}, ClArTTS~\cite{kulkarni2023clartts}, ASC \cite{halabi2016arabic}, and Common Voice~\cite{ardila2020common}, SADA \cite{10446243}, and others. We also used MADAR~\cite{bouamor-etal-2018-madar} and NADI~\cite{abdul-mageed-etal-2023-nadi} text datasets for pretraining.
In our experiments, we compare the following:

\begin{itemize}
\setlength{\parskip}{0pt}
\item \textbf{v1}: This variant is as described in \citet{toyin-etal-2023-artst}, pretrained only on MSA. 
\item \textbf{v2}: In this variant, we use a mixture of MSA and dialectal data in pretraining. 
\item \textbf{v3}: In this variant, we use a mixture of MSA, dialectal, and multilingual data in pretraining. 
\end{itemize}

See Table \ref{tab:dataset_summary_combined_pretraining} in Appendix~\ref{app:data} for details of all the datasets used in pre-training.

\begin{table}
     \centering
     \resizebox{\columnwidth}{!}{
     \begin{tabular}{llrr}
        \toprule
        \textbf{Dataset} & \textbf{Dialect} & \textbf{Hours} & \textbf{Words} \\
        \midrule
        QASR    & MSA & 2000 hrs & 13.33 M \\
        \midrule
        MGB2    & MSA & 1000 hrs & 7.31 M \\
        \midrule
        MGB3$\texttt{[ASR]}$    & EGY        & 2.83 hrs   & 18.93 K \\
        \midrule
        MGB5$\texttt{[ASR]}$    & MOR        & 6.74 hrs   & 56.97 K \\
        \midrule
        SADA~\cite{10446243}    & SAU        & 418 hrs  & 3.25 M \\
        \midrule
        Mixat~\cite{al-ali-aldarmaki-2024-mixat}  & UAE & 15 hrs & 57.94 K \\
        \midrule
        TARIC-SLU~\cite{mdhaffar-etal-2024-taric}& TUN & 8 hrs & 72.00 K \\
        \midrule
        ParallelCorp~\cite{6487288}& MSA & 32 hrs & 30.66K \\
                     & GLF & 32 hrs & 27.26K \\
                     & LEV & 32 hrs & 18.43K \\
                     & EGY & 32 hrs & 48.56K \\
        \midrule
        MASC~\cite{e1qb-jv46-21} & MSA & 612.28 hrs & 3.80 M \\
             & SAU & 452.24 hrs & 301.92 K \\
             & SYR & 211.33 hrs & 1.06 M \\
             & EGY & 175.36 hrs & 1.03 M \\
             & JOR & 42.21 hrs  & 330.83 K \\
             & LEB & 25.20 hrs  & 155.76 K \\
             & IRA & 17.37 hrs  & 121.12 K \\
             & TUN & 12.17 hrs  & 34.34 K \\
             & Multiple & 10.57 hrs  & 80.08 K \\
             & UAE & 9.87 hrs   & 6.42 K \\
             & MOR & 8.60 hrs   & 58.38 K \\
             & PAL & 6.17 hrs   & 45.35 K \\
             & KUW & 4.04 hrs   & 32.37 K \\
        \bottomrule
     \end{tabular}}
     \caption{Summary of Dataset Statistics for \textbf{Fine-Tuning}: Hours of Audio, Word Counts, and Associated Dialects. Multiple is mix of several dialects not neccessary from the listed dialects (no information from the source).}
     \label{tab:dataset_summary_combined_finetuning}
 \end{table}

\subsection{Dialectal Fine-Tuning}

The datasets we use for dialectal fine-tuning are shown in Table~\ref{tab:dataset_summary_combined_finetuning}. We collected as many open-source data as needed to maximize coverage of dialects. Note that, for MGB5 and MGB3, as the data is based on YouTube videos, many of the originally referenced videos are no longer available, so at the time of our experiments, only 2.5 hours of training were available for MGB3 and 2 hours for MGB5. Furthermore, multi dialectal datasets, such as MASC~\cite{e1qb-jv46-21}, have unbalanced representation of dialects. 
The high-resource dialects in our collection include SAU, SYR, EGY, and MSA; each has at least 200 hours of transcribed ASR data. 
UAE, MOR, JOR, LEB, IRA, and TUN have a medium amount of fine-tuning data between 10 and 50 hours. KUW and PAL are low-resource dialects with less than 10 hours of transcribed data in total. Finally, we left ALG, YEM, and SUD from the MASC dataset for zero-shot testing.

Figure \ref{fig:dialect_distribution_stata} illustrates the distribution of dialectal data we use for pre-training and fine-tuning our dialectal model. We exclude MSA from the figure as it has disproportionally more data than all dialects.

\subsection{Multi-Lingual Fine-Tuning}
In addition to the above dialectal data, we used English, French, and Spanish sets for multi-lingual fine-tuning and code-switching experiments described in section \ref{sec:cs}. English and French are commonly spoken in various Arabic-speaking countries, and to a lesser extent, Spanish is spoken in some parts of North Africa. More details about the datasets used in these experiments are provided in section \ref{sec:cs}.

\subsection{Experimental Settings}

For partitioning the data into training, development, and test sets, we adhered to the official splits provided with each dataset.
We followed the data preparation and training methodology established in the original ArTST implementation. For comprehensive details regarding the model architecture and data preprocessing, readers are directed to \citet{toyin-etal-2023-artst}. 

\begin{figure}
    \centering
    \includegraphics[width=\linewidth]{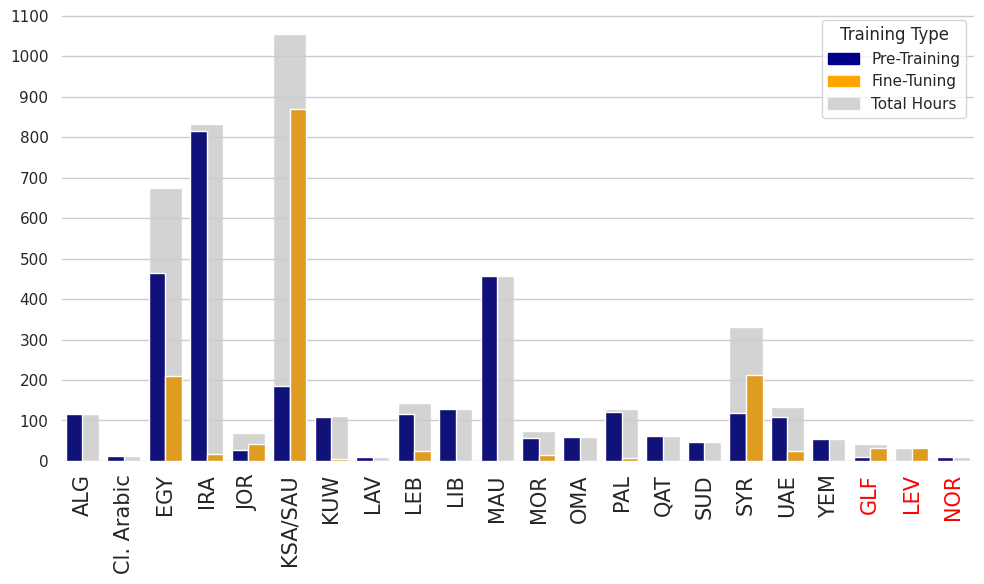}
    \caption{Distribution of dialectal speech data in pre-training and fine-tuning. MSA 
    data are not shown.}
    \label{fig:dialect_distribution_stata}
\end{figure}


\paragraph{Computational Details} The pre-training was executed on a cluster of 4 A100 GPUs over a duration of 14 to 21 days for each model. We used Adam optimizer with a learning rate of $2\times 10^{-4}$, spanning 335K updates, and a warm-up phase of 64K updates. The maximum speech token length was set at 250K (equivalent to 15.625 seconds). Each fine-tuning experiment was run on one A100 GPU over a duration of 7 days (MGB2, QASR, MASC, SADA) or 2 days for smaller sets (MGB3, MGB5, etc.). We used Adam optimizer with a tri-stage scheduler with learning rate of $6\times 10^{-5}$. The total computational budget for all experiments is estimated to be $\sim$6000 GPU-hours.

\paragraph{Normalization}
Prior to model training, we implemented the same data normalization steps outlined in \citet{toyin-etal-2023-artst}. In addition, we applied post-prediction normalization steps before calculating Word Error Rates (WER), following standard practices in Arabic ASR. All reported results reflect post-normalization performance. The normalization script, sourced from a publicly available GitHub repository\footnote{\url{github.com/iamjazzar/arabic-nlp/blob/master/normalization/orthographic_normalization.py}}, performs orthographic standardization of \textit{Alef}, \textit{Yaa}, and \textit{Taa} characters.

\section{Effect of Pre-Training Data}
We first examine the effect of pre-training on downstream ASR performance. As described in section \ref{sec:pre-train-data}, we compare a model pre-trained mainly on MSA (ArTST \textbf{v1}), and our multi-dialectal version (henceforth \textbf{v2}). Note that pre-training does not utilize aligned speech and text; it incorporates unaligned speech and text data for self-supervised learning. For these experiments, we use the same finetuning data, and only vary the pretraining sets. 

\subsection{Benchmarking MSA Performance}
We first report results on benchmark datasets to compare the performance of both models against the state-of-the-art. 
MGB2 is the main benchmark for MSA speech recognition. We evaluated the performance of ArTST v1 and v2 fine-tuned in MGB2, compared to existing SOTA models, in Table~\ref{tab:experiment_results_mgb2}. The results show that incorporating dialectal data in pretraining does not negatively affect MSA performance, as we achieve the best WER of 12.48\% and 12.39\%, with and without LM fusion.

\begin{table}
        \centering
        \resizebox{\columnwidth}{!}{
        \begin{tabular}{lcc}
        \toprule
             \textbf{System} & \textbf{WER(\%)} & \textbf{CER(\%)} \\
        \midrule
            \multicolumn{3}{l}{From \cite{hussein2022arabic}:} \\
            HMM-DNN                     & 15.80 & --- \\
            E2E, CTC + LM               & 16.90 & --- \\
            E2E, Attention + LM         & 13.40 & --- \\
            E2E, CTC , Attention + LM   & 12.50 & --- \\
        \midrule
            ArTST v1 + LM \cite{toyin-etal-2023-artst} & 12.78 & 6.33 \\
        \midrule
            v2       & 12.49 & 6.44 \\
            v2 + LM               & \textbf{12.39}          & 6.51 \\
        \bottomrule
        \end{tabular}}
        \caption{Comparing our performance against models reported in \citet{hussein2022arabic} and \citet{toyin-etal-2023-artst}, which include the best performing models previously reported on MGB2.}
        \label{tab:experiment_results_mgb2}
    \end{table}

\subsection{Benchmarking Dialectal Performance}
Tables~\ref{tab:experiment_results_mgb3} and~\ref{tab:experiment_results_mgb5} show the performance of the models on the dialectal MGB3 (Egyptian) and MGB5 (Moroccan) benchmarks. Each of these benchmarks contain multiple references as dialectal speech has no standard spelling. We report the average and multi-refrence WER for our model variants, and compare against the best model in each challenge, as well as the SOTA model in each benchmark. Each model is first fine-tuned on MSA, then fine-tuned again on the target MGB train sets.  We also report the results of the large multilingual models: Whisper~\cite{radford2023robust} and MMS~\cite{pratap2024scaling}, fine-tuned on the same set.  We refer to the Arabic data the models are previously fine-tuned on as \textit{`Adaptation'} data. Starting with MSA data before fine-tuning on the target dialect has previously been established as an effective strategy for dialectal ASR \cite{ali2017speech}.

\begin{table}
    \centering
    \resizebox{\columnwidth}{!}{
    \begin{tabular}{lcccccc}
    \toprule
    \bf System & \bf Adaptation & \bf Fine-Tuning & \bf AV-WER & \bf MR-WER \\
    \midrule
    \textbf{Aalto}  &   MGB2    &  MGB3      &   37.50 & 29.30\\
    \midrule
    \multirow{3}{*}{\textbf{Whisper}}    & ComVoice  & \multirow{3}{*}{MGB3}  & \multirow{3}{*}{39.04}    & \multirow{3}{*}{24.92}\\
                                & Fleurs   \\ 
                                & Covost2  \\
    \midrule
    \multirow{2}{*}{\textbf{MMS}}        & BibleTrans & \multirow{2}{*}{MGB3}  & \multirow{2}{*}{100.04}    & \multirow{2}{*}{99.92} \\
                                & NewTestamentRec \\
    \midrule
                \textbf{v1}  & MGB2  & MGB3  & 37.08 & 29.39 \\
    \midrule
                \textbf{v2}  & MGB2  & MGB3  &  \textbf{33.20} & \textbf{25.28} \\
    \bottomrule
    \end{tabular}}
    \caption{WER(\%) on MGB3 Egyptian ASR. Aalto is the best system in the MGB3 challenge \cite{ali2017speech}}
    \label{tab:experiment_results_mgb3}
    
    \vspace{1em}
    
    \resizebox{\columnwidth}{!}{
    \begin{tabular}{lcccccc}
    \toprule
    \bf System & \bf Adaptation & \bf Fine-Tuning & \bf AV-WER & \bf MR-WER \\
    \midrule
    \textbf{RDI-CU}             & MGB2      & MGB5     & 59.40     & 37.60 \\
    \midrule
    \multirow{3}{*}{\textbf{Whisper}}    & ComVoice  & \multirow{3}{*}{MGB5}  & \multirow{3}{*}{164.13}    & \multirow{3}{*}{227.34}\\
                                & Fleurs   \\ 
                                & Covost2  \\
    \midrule
    \multirow{2}{*}{\textbf{MMS}}        & BibleTrans & \multirow{2}{*}{MGB5}  & \multirow{2}{*}{111.89}    & \multirow{2}{*}{102.30} \\
                                & NewTestamentRec \\
    \midrule
            \textbf{v1}                  & MGB2  & MGB5  &  49.39    & \textbf{27.95} \\
    \midrule
            \textbf{v2}      & MGB2   & MGB5  & \textbf{48.91}     & 28.02 \\
    \bottomrule
    \end{tabular}}
    \caption{WER(\%) on Moroccan ASR. RDI-CU is the best system in the MGB5 challenge \cite{ali2019mgb}}
    \label{tab:experiment_results_mgb5}
\end{table}

\begin{table*}[h]
\centering
\scalebox{0.8}{
\begin{tabular}{l|cccc|cc}
\toprule
\multirow{2}{*}{\textbf{Dataset}} & \multicolumn{4}{c}{\textbf{Zero-Shot}} & \multicolumn{2}{|c}{\textbf{Fine-Tuning}} \\ 
 & \textbf{Whisper} & \textbf{MMS} & \textbf{v1} & \textbf{v2} & \textbf{v1} & \textbf{v2}  \\ 
\hline
\textbf{TARIC-SLU} \hspace{0.85em}(TUN)  & 138.14 & 93.54 & 107.56 & 106.46   & \textbf{14.70} & 14.80 \\ 
\hline
\textbf{ParallelCorp} (MULT)  & 99.17  & 83.16 & 128.72 & 141.92   & 9.57 & \textbf{9.31} \\ 
\hline
\textbf{SADA} \hspace{1em}(SAU)& 82.16 & 78.28 & 39.41 & \textbf{29.77}       & 39.24 & 29.91 \\ 
\hline
\textbf{MASC}        &  &  &  &  &  &  \\
SAU & 48.39 & 65.30 & 61.23 & 58.72           & 27.40           &  \textbf{27.33}  \\
SYR & 26.65 & 33.21 & 21.99 & 18.37           & 18.64           &   \textbf{17.42} \\
EGY & 41.73 & 66.04 & 50.87 & 47.17           & 38.47           &  \textbf{36.43}  \\
JOR & 28.65 & 54.63 & 61.23 & 34.97           & \textbf{19.72} & 21.08   \\
LEB & 40.95 & 64.58 & 35.65 & 42.66           & 30.01 &  \textbf{28.05}  \\
IRA & 41.69 & 59.33 & 50.46 & 48.03           & \textbf{31.10} &  34.64  \\
TUN & 47.45 & 60.58 & 50.37 & 46.67           & 19.26 &  \textbf{18.52}  \\
MOR & 65.87 & 80.84 & 78.92 & 66.87           & \textbf{47.59} &  49.40  \\
PAL & \textbf{53.20} & 83.72 & 77.94 & 73.53  & 55.88 &  53.53  \\
KUW & \textbf{36.00} & 81.71 & 64.74 & 52.02  & 50.29 &  46.24  \\
\bottomrule
\end{tabular}
}
\caption{WER (\%) in zero-shot and fine-tuning settings. We compare zero-shot performance of Whisper, MMS, ArTST v1, and Our dialectal pretraining (v2). ArTST v1 and v2 are finetuned on MGB2 (MSA), whereas Whisper and MMS are finetuned/pretrained with multi-lingual data, including Arabic.}
\label{tab:results_on_dataset_collection}
\end{table*}



In MGB3, dialectal pretraining (v2) results in about 4\% absolute reduction in WER, establishing a new SOTA result on this benchmark. Smaller improvement in terms of Average WER is observed for MGB5, where there is no clear advantage observed using our dialectal version. 
This difference is likely a result of our pre-training having a lot more Egyptian than Moroccan data (see Figure \ref{fig:dialect_distribution_stata}).

\subsection{Zero-Shot \& Fine-Tuning Results}
To further quantify the effect of dialectal pre-training, we evaluate the performance of our model across different datasets. We first fine-tune models on MSA using MGB2 dataset. We test the model performance on dialects directly (zero-shot) and with dialectal fine-tuning. The results are shown in Table \ref{tab:results_on_dataset_collection}.
On average, we see improvements in performance in both zero-shot and fine-tuning experiments using dialectal pretraining (v2) compared to MSA-centric pretraining (v1). We also see that both models perform better than Whisper and MMS in zero-shot settings in most cases. There are some exceptions, such as in KUW, where Whisper performs better than all other models, including the fine-tuned models, but in most cases v2 performs best. This underscores the advantage of monolingual models compared to multilingual performance, as observed in \citet{toyin-etal-2023-artst} and \citet{radford2023robust}. In addition, the results underscore the importance of dialectal coverage in pretraining: the cases where v2 performs worse than v1 are all dialects for which pretraining data are limited, such as TUN (no pretraining data) and JOR (smallest dialect size in pretraining). 

\section{Joint Models \& Dialect ID}
\label{sec:data_augmentation_dialect_id}
So far, models were first fine-tuned on MSA, followed by additional fine-tuning on each target dialect. This results in a separate model per dialect, which incurs memory costs and may have practical limitations as it requires advance knowledge of the dialect ID for deploying the correct model. 

In this section, we assess the relative effectiveness of individual dialectal fine-tuning compared with joint dialect fine-tuning, where we train a single model for all dialects. To that end, we joined multiple dialectal train sets from MASC, as shown in Table \ref{tab:joint_dialect_training_stata}.
We excluded ALG, YEM, SUD for zero-shot evaluation. The resulting joint corpus consists of 12 dialects including MSA, with approximately 1,501 hours in total. We fine-tuned a single joint model using this data.

\begin{table}[h]
    \centering
    \resizebox{0.9\columnwidth}{!}{
    \begin{tabular}{lrrc}
    \toprule
    \textbf{Dialect} & \textbf{Hours} & \textbf{Words} & \textbf{Source} \\
    \midrule
        MSA & 612.28 hrs & 3.80 M   & MASC\\
        SAU & 452.24 hrs & 301.92 K & SADA, MASC\\
        SYR & 211.33 hrs & 1.06 M   & MASC \\
        EGY & 175.36 hrs & 1.03 M   & MGB3, MASC\\
        JOR & 42.21 hrs  & 330.83 K & MASC \\
        LEB & 25.20 hrs  & 155.76 K & MASC \\
        IRA & 17.37 hrs  & 121.12 K & MASC \\
        TUN & 12.17 hrs  & 34.34 K  & TARIC-SLU, MASC \\
        UAE & 9.87 hrs   & 6.42 K   & Mixat, MASC\\
        MOR & 8.60 hrs   & 58.38 K  & MASC \\
        PAL & 6.17 hrs   & 45.35 K  & MASC \\
        KUW & 4.04 hrs   & 32.37 K  & MASC \\
    \bottomrule
    \end{tabular}} 
    \caption{Datasets used to train the joint models.}
    \label{tab:joint_dialect_training_stata}
\end{table}

\begin{table*}
\centering
\resizebox{1.75\columnwidth}{!}{%
\begin{tabular}{l|cc|c|ccc}
\toprule
\textbf{Approach} & \multicolumn{2}{c|}{\textbf{Zero-Shot}}  & \textbf{Fine-Tuning} & \textbf{No Dialect ID} & \textbf{Dialect Forcing} & \textbf{Dialect Inference}\\
\hline
\textbf{Fine-Tuning Data} & \textbf{\textit{MGB2}} & \textbf{\textit{QASR}} & \textbf{\textit{MGB2$\rightarrow$ Target}} & 
\multicolumn{3}{c}{\textbf{\textit{Joint Mutli-Dialectal Set (Table \ref{tab:joint_dialect_training_stata})}}}\\
\hline
    SAU & 58.72 & 43.41  & \textbf{27.33}  & 29.41 & 30.56 & 29.94 \\
    SYR & 18.37 & \textbf{16.20} & 17.42  & 19.20 & 22.41 & 20.30 \\
    EGY & 47.17 & \textbf{38.78} & 36.43  & 45.17 & 61.06 & 46.79 \\
    JOR & 34.97 & 25.42          & 21.08  & \textbf{19.63} & 21.49 & 20.11 \\
    LEB & 42.66 & 40.51          & 28.05  & 28.22 & 29.43 & \textbf{26.89} \\
    IRA & 48.03 & 40.27          & 36.10  & \textbf{29.33} & 31.75 & 30.83 \\
    TUN & 46.67 & 45.93          & \textbf{26.67}  & 37.23 & 28.47 & 27.74 \\
    MOR & 66.87 & 55.42          & 56.63 & 57.49 & 53.89 & \textbf{49.10} \\
    PAL & 73.53 & 45.59          & 53.53  & 46.22 & \textbf{43.90} & 44.48 \\
    KUW & 52.02 & 45.09          & 46.24  & \textbf{35.43} & 39.43 & 37.71 \\
    MSA & 21.09 & 16.78        & 15.34  & \textbf{11.59} & 12.66 & 12.09\\
    \hline
    Macro Average & 46.37 & 37.58 & 33.17 & 32.63 & 34.09 & \textbf{31.45}  \\
\bottomrule
\end{tabular}

}
\caption{WER (\%) of various models compared with joint dialectal fine-tuning with different dialect ID strategies.}
\label{tab:expanded-dialect-error-rates}
\end{table*}

\subsection{Dialect ID}
We trained another model with the aforementioned joint dataset, but with the inclusion of explicit dialect identifiers. We augmented the dictionary with special tokens for dialect IDs, and used them to prepend the decoding string: 

\(
\texttt{
<S> DIALECT T$_{1}$ T$_{2}$ ... T$_{n}$ </S>
}
\)

For inference, we experimented with two strategies: (1) \textbf{Transcribing with dialect forcing}, where we manually add the dialect ID to condition the decoder output; the decoder is forced to start predictions with the tokens  \verb|<S> DIALECT| . (2) \textbf{Transcribing with dialect inference}, where we let the model predict the dialect token. We use this approach for zero-shot ASR on unseen dialects  (Table \ref{tab:zero_shot_dialects_performance}).

The results of the models trained with joint dialects compared to models trained on MGB2 and QASR are shown in Table \ref{tab:expanded-dialect-error-rates}. Note that both MGB2 and QASR contain mostly MSA, but also a small amounts of various dialects, but their exact distribution is unknown. We also reproduce the fine-tuning results from Table \ref{tab:results_on_dataset_collection} for easy comparison. We see that joint modeling results in improvement for low-resource dialects, including: JOR, TUN, and KUW, but degrades performance of the high-resource SYR and EGY dialects. Interestingly, dialect forcing was worse on average than joint modeling with no dialect ID, while dialect inference resulted in the best performance overall. We surmise that the model learns dialectal patterns that do not perfectly align with the dialect ID as indicated in the training data. Since the dialect IDs are coarse country-level approximations, letting the model infer the dialect based on the speech is the best approach for most cases. Many dialectal sets, such as SYR and SAU, contain a lot of MSA utterances that are incorrectly identified as dialectal. 

\begin{figure}[h]
\centering
\includegraphics[width=0.46\textwidth]{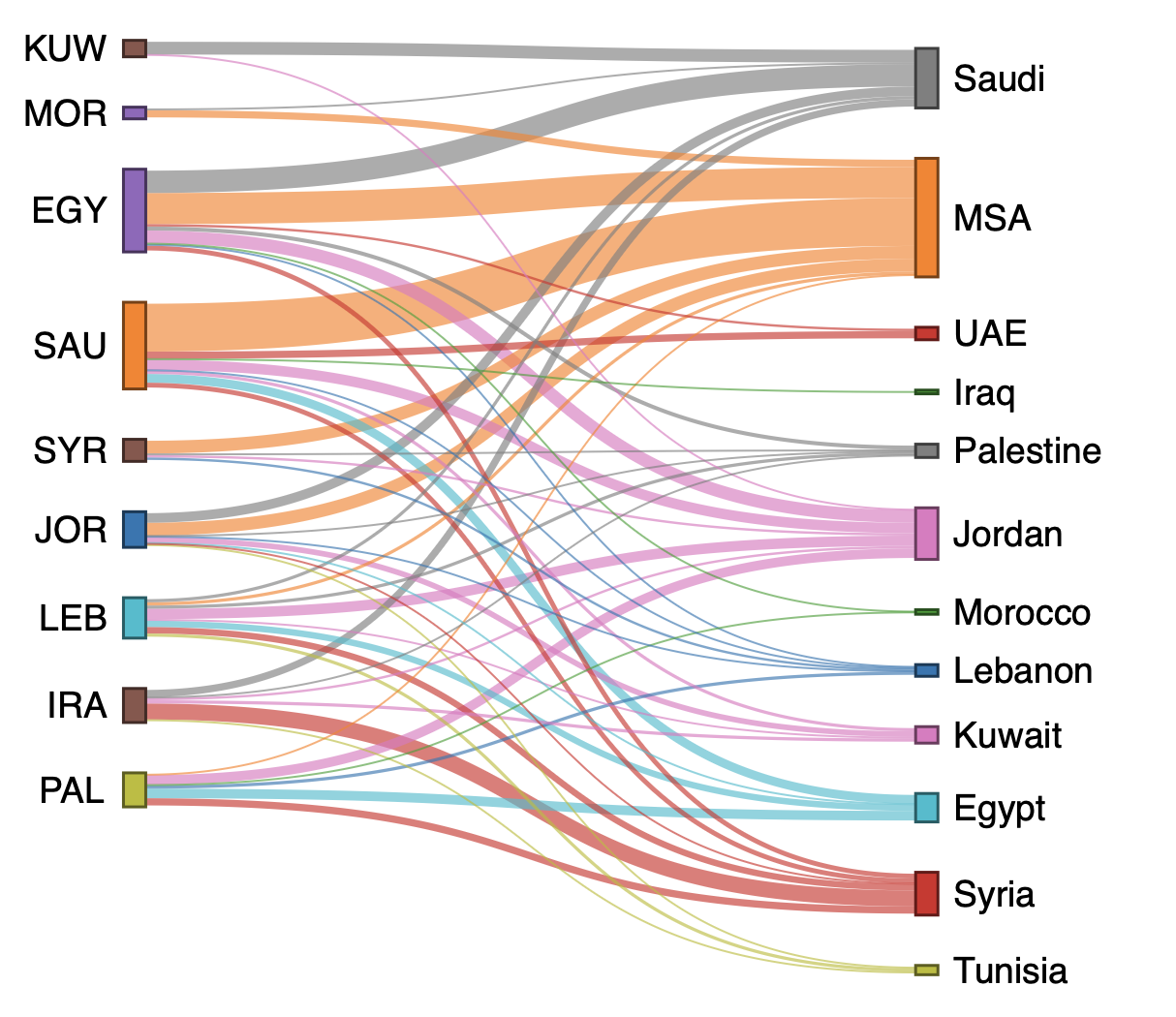}
\caption{Dialect identification performance: true (left), predicted (right). All lines are proportional to their ratio over the total errors except for SAU$\rightarrow$MSA, which is reduced 5 times for clarity.}
\label{fig:DID}
\end{figure}

Figure \ref{fig:DID} illustrates dialect inference errors. Note that the number of errors is proportional to the test data size. The overall dialect identification performance is around 90\%. Some low-resource dialects, such as KUW, are predicted as their closest high-resource variant, such as SAU, resulting in worse performance compared to joint models without a dialect ID, but on average, dialect forcing leads to the lowest WER. 

\section{Effect of Data Adaptation}

\begin{table}
    \centering
    \resizebox{0.9\columnwidth}{!}{
    \begin{tabular}{l|ccc}
    \toprule
    \diagbox{\textbf{Fine-Tuning}}{\textbf{Adaptation}} & --- & MGB2 & QASR \\
    \midrule
    MGB3 & 136.52 & 25.28 & \textbf{19.53} \\
    MGB5 & 94.84 & 28.02 & \textbf{27.58} \\
    TARIC-SLU & 30.46 & 14.80 & \textbf{14.48} \\
    ParallelCorp & 28.97 & 9.31 & \textbf{9.08} \\
    SADA & 29.77 & 29.91 & \textbf{29.55} \\
    Mixat & 100.0 & \textbf{33.40} & 35.21 \\
    \bottomrule
    \end{tabular}}
    \caption{WER (\%) of fine-tuned models on various datasets with different adaptation methods: ---: no adaptation, MGB2, or QASR. }
\end{table}

In the above experiments, we followed the strategy of initializing the models by first fine-tuning on MSA data. In most cases, we used MGB2 as the base model, following previously established results on Egyptian ASR \cite{ali2017speech}. This adaptation approach is meant to enhance the performance on low-resource dialects, facilitating faster convergence with limited training samples. 
However, as pre-training on more diverse sets proved to be effective, adaptation on more diverse data is also likely to be fruitful. As observed in Table \ref{tab:expanded-dialect-error-rates}, models trained on QASR resulted in far better zero-shot performance, approaching the performance of joint-dialects models. This is attributed to the fact that QASR is both larger in size and known to have more dialectal data compared to MGB2 (but both have no documented statistics of dialectal coverage). To validate this observation, we experimented with dialectal fine-tuning adapted from two variants:  one based on MGB2 and one based on QASR~\cite{mubarak-etal-2021-qasr}, followed by dialect-specific fine-tuning. Table \ref{tab:experiment_results_mgb5} shows fine-tuning results with no adaptation (directly fine-tuning on the target dialect), compared with starting from MGB2 or QASR. First, our results corroborate the previous findings that adapting models from MSA results in large reduction in error rates. In all except the Mixat dataset, starting from QASR results in better performance compared to MGB2. However, the difference is negligible except on MGB3 Egyptian set (around 6\% absolute WER reduction). There are two factors that we speculate  underline this result: The small size of the MGB3 set, and the existence of Egyptian dialect in the QASR corpus more substantially than the other dialects. Overall, using the QASR dataset as a basis for adapting dialectal models is recommended as it improves or maintains performance. 

\section{Zero-Shot Performance}

\begin{table}
\centering
    \resizebox{\columnwidth}{!}{
    \begin{tabular}{l|ccc}
        \toprule 
        \multirow{2}{*}{\diagbox{\textbf{System}}{\textbf{Dialect}}}
                                        & \textbf{ALG}  & \textbf{SUD}   & \textbf{YEM} \\
                                        &&&\\
        \hline
        \textbf{ArTST v1}$\rightarrow$MGB2                 &  73.18  & 69.20  & 41.64 \\
        \hline
        \textbf{v2}$\rightarrow$MGB2          &  70.82  & 69.31  & 39.45  \\
        \textbf{v2}$\rightarrow$QASR               &  51.72  & 46.64 &  34.78 \\
        \hline
        \textbf{v2} $\rightarrow$Joint 
        &  \textbf{45.20}  &  40.69 & 33.08 \\
        \textbf{v2} $\rightarrow$Joint (w. dialect inference)
        &  47.12  &  \textbf{40.15} & \textbf{31.84} \\
        
        \bottomrule
    \end{tabular}}
    \caption{WER\% of various models on held-out dialects.}
    \label{tab:zero_shot_dialects_performance}
\end{table}
We show the zero-shot performance on the three held-out sets: ALG, SUD, and YEM. We compare the baseline, v1, with multi-dialectal pretraining (v2). We also compare models fine-tuned on MGB2, QASR, or our joint dialectal set. The results are shown in Table \ref{tab:zero_shot_dialects_performance}. 
Our model achieves slightly lower error rates compared to v1, even when fine-tuned on the same MGB2 set. Better performance is achieved with QASR, which includes some dialectal data. The joint dialectal fine-tuning achieves the best performance on the held-out dialects.  In general, performance in held-out sets is on a par with low-resource dialects, with WER above 30\%. Table \ref{tab:zero_shot_cx_dialects_performance} in the Appendix shows the zero-shot performance after fine-tuning on a single target dialect.

\section{Code-Switching Performance}\label{sec:cs}

The models analyzed so far were trained exclusively on Arabic data. While small amounts of code-switching (CS) exist in these sources, they are insufficient to learn the characteristics of the embedded languages. Large multi-lingual models are generally more effective on CS data~\cite{kadaoui-etal-2024-polywer}, even if they are less competent on monolingual Arabic. To make our models more inclusive, improving performance in the presence of code-switching is necessary. 
To that effect, we train a multilingual version of the model (we will refer to this as \textbf{v3}). The pre-training data for this version are listed in Table \ref{tab:dataset_summary_combined_pretraining} in the Appendix.
We test \textbf{v3} against \textbf{v1} and \textbf{v2} on 
available CS data for Arabic: ArZN \cite{al2024arzen} for Egyptian-English speech, Mixat \cite{al-ali-aldarmaki-2024-mixat} for Emirati-English speech, and TunSwich \cite{abdallah2024leveraging} for Tunisian-French speech.
We also train a joint multi-lingual model (without dialect or language ID). In addition to the datasets described in Table \ref{tab:joint_dialect_training_stata}, we add the multi-lingual and code-switching data shown in Table \ref{tab:joint_multilingual_data}. 

In Table \ref{tab:results_cs}, we show the performance of models finetuned directly from the pretrained checkpoints, or finetuned from existing ASR checkpoints (MGB2 checkpoint for v1, joint multi-dialectal checkpoint for v2, and joint multi-lingual checkpoint for v3). 
First, for models fine-tuned directly on the target set, we observe that multilingual pretraining significantly improves performance across all CS test sets, resulting in more than 10\% absolute reductions in WER for all test sets. This clearly illustrates the advantage of multi-lingual pretraining in code-switching scenarios.  
We also evaluated models  initialized from the joint models followed by target fine-tuning on the CS train sets, and this reduced error rates further. The best performing model is the joint multilingual v3 mode, with 4 to 7\% absolute reduction in WER compared to the second best model.  
We show examples of ASR outputs from the three CS datasets using the various models in Figure~\ref{fig:code-switching-examples} in the Appendix.

\begin{table}
    \centering
    \resizebox{0.8\columnwidth}{!}{
    \begin{tabular}{lrrc}
    \toprule
    \textbf{Languages} & \textbf{Hours} & \textbf{Words} & \textbf{Source} \\
    \midrule
        EN & 1601.92 hrs & 10.35 M   & CommonVoice\\
        FR & 732.02 hrs & 5.03 M & CommonVoice\\
        SP & 408.34 hrs & 2.79 M   & CommonVoice \\
        TUN-FR & 10.89 hrs & 70.86 K   & TunSwitch\\
        UAE-EN & 8.97 hrs  & 57.82 K & Mixat \\
        EGY-EN & 5.61 hrs  & 52.00 K & ArzEn \\
    \bottomrule
    \end{tabular}} 
    \caption{Additional datasets used to train the joint multilingual model.}
    \label{tab:joint_multilingual_data}
\end{table}


\begin{table}
\centering
\scalebox{0.61}{



\begin{tabular}{lccccccc}
\toprule
\textbf{Adaptation data} & - & - & - & MSA & Dialectal & Multilingual \\
\midrule
\textbf{Test Set} & 
\textbf{v1} & \textbf{v2} & \textbf{v3} & \textbf{v1} & \textbf{v2} & \textbf{v3}\\
\midrule
MGB2                    
& 13.42   & \textbf{12.5}    & 13.0       & - & - &  - \\
\midrule
ArzEn                   & 
43.21   & 77.59            & 35.26      & 34.85 & 33.71 & \textbf{27.43} \\
TunSwitch               & 
53.85   & 101.94           & 40.68      & 43.87 & 43.59 & \textbf{36.66} \\ 
Mixat                   & 
42.50   & 92.41            & 34.27      & 27.07 & 25.73 & \textbf{21.66} \\
\bottomrule
\end{tabular}
}
\caption{ASR Results using the various checkpoints: \textbf{v1}, \textbf{v2} and \textbf{v3}. We compare models trained directly from the pretrained checkpoint vs. starting with an ASR checkpoint trained with the specified \textbf{adaptation data}: 
MSA adaptation using the MGB2 dataset; Dialectal adaptation using data listed in Table~\ref{tab:joint_dialect_training_stata}; Multilingual adaptation using data from Table~\ref{tab:joint_dialect_training_stata} and Table \ref{tab:joint_multilingual_data}.
}
\label{tab:results_cs}
\end{table}

\begin{figure}[ht]
    \centering
    \hspace{-0.4cm}\includegraphics[width=1.05\linewidth]{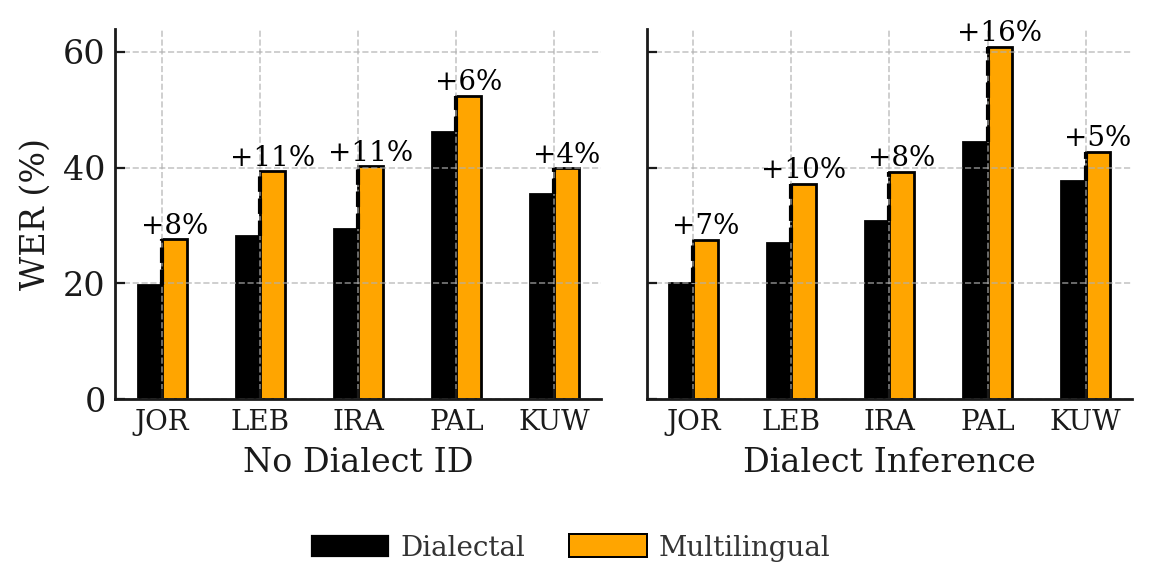} 
    \caption{WER (\%) and absolute difference on a subset of dialects, comparing our joint dialectal fine-tuning vs. joint multi-lingual fine-tuning on Arabic dialects.}
    \label{fig:wer_comparison}
\end{figure}

\paragraph{Language Interference:} We test the effect of multi-lingual pre-training on MSA performance. Language interference is known to negatively affect monolingual performance \cite{toyin-etal-2023-artst},  so we test our multilingual model on the MGB2 benchmark to quantify this effect (see Table \ref{tab:results_cs}). The model achieves 13.0\% WER, which is indeed worse than the SOTA result we achieve with the Arabic-only model (see Table \ref{tab:experiment_results_mgb2}), but the difference at 0.5\% absolute WER is rather small.
When it comes to dialects, however, we find that language interference  has a significant negative effect, resulting in 4\% to 16\% absolute increase in error rates, as shown in Figure \ref{fig:wer_comparison}. 

\section{Conclusions}
We presented the largest study on dialectal Arabic ASR to empirically demonstrate the effect of various training paradigms on ASR performance. We compared models pre-trained with and without dialects, in high, low, and medium-resource settings, in addition to zero-shot. We find that overall, \textbf{dialectal pre-training improves performance in zero-shot and low-resource cases}, and mostly maintains performance on MSA and high-resource dialects. We also find that all dialects benefit from adaptation of models pre-fine-tuned on MSA, and this effect is most noticeable for low and medium-resource dialects. 
We experimented with  multi-dialectal fine-tuning, where we joined the train sets of 12 dialects. We observe performance improvements 
on average, and at least the same performance as the target-dialect fine-tuning setting, and the best performance on held-out dialects. Interestingly, while using dialect ID in decoding is effective, \textbf{forcing the dialect ID results in worse performance compared to dialect inference}. While joint training results in improved performance for the medium and low-resource dialects, \textbf{target-dialect fine-tuning is more effective for high-resource dialects}. Finally, we experimented with multi-lingual pre-training and fine-tuning for improving performance on code-switched utterances, and achieved significant reductions in error rates on all available test sets. However, reductions in monolingual performance were also observed due to language interference, particularly for dialects, were WER increased by 4\% to 16\% for some dialects.  
To enable easier adoption and further experiments, we released the pretrained dialectal and multilingual checkpoints, the fine-tuned MGB2 models, and the joint dialectal and multilingual ASR models. 
\section*{Limitations}
One of the limitations in dialect-related work is the coarse classification of dialect IDs; dialects in our datasets are classified by regions or countries, whereas actual dialectal variations are far more fine-grained. For example, the Saudi dataset, SADA, covers a large geographical area and many dialects, but it is considered as one dialect based on our classification. Moreover, the way the datasets are collected do not guarantee that the data are indeed dialectal. For instance, with manual inspection of the Syrian test and dev sets from MASC, we observed that all instances are in MSA rather than Syrian dialects. In addition, Arabic dialects are spoken varieties that do not have a standard spelling system. This results in large variations in transcriptions, but standard WER does not account for these variations, resulting in more pessimistic results. With the exception of the MGB3 and MGB5 benchmarks where we report average and multi-reference WER across 4 references, all datasets have only a single reference.

\bibliography{template}

\appendix
\clearpage

\section{Pre-Training Dataset Statistics} \label{app:data}
Table \ref{tab:dataset_summary_combined_pretraining} shows the complete list of datasets used for pre-training v1, v2, and v3. 
 \begin{table}[H]
     \centering
     \hspace{2pt}
     \scalebox{0.63}{
     \hspace{-5pt}\begin{tabular}{l|c|rrccc}
        \toprule
        \textbf{Dataset} & \textbf{Dialect} & \textbf{Hours} & \textbf{Words} & \textbf{v1} & \textbf{v2} &\textbf{v3} \\
        \hline
        ASC                          & MSA               & 3.7 hrs      & 20.58 K   &               & $\checkmark$ & $\checkmark$ \\ \hline
        ArzEn$^{\texttt{[cs]}}$      & EGY               & 5.61 hrs      & 52.00 K   &               &  & $\checkmark$ \\ \hline
        CommonVoice                  & Dialect Mix               & 133.24 hrs   & 494.83 K  &               & $\checkmark$ & $\checkmark$ \\ \hline
        \multirow{3}{*}{CommonVoice} & ENG               & 1601.92      & 10.35 M           &               &            & $\checkmark$ \\ 
                                     &      FR           & 732.02       & 5.03 M           &               &            & $\checkmark$ \\ 
                                     &          ES       & 408.34       & 2.79 M           &               &            & $\checkmark$ \\ \hline
        ClArTTS                      & CA                & 12 hrs       & 76.31 K   &               & $\checkmark$ & $\checkmark$ \\ \hline
        \multirow{12}{*}{MASC}       & EGY               & 175.36 hrs   & 1.03 M    &               &            & $\checkmark$\\ 
                                     & IRA               & 17.37 hrs    & 121.12 K  &               &            & $\checkmark$\\ 
                                     & JOR               & 42.21 hrs    & 330.83 K  &               &            & $\checkmark$\\ 
                                     & KUW               & 4.04 hrs     & 32.37 K   &               &            & $\checkmark$\\ 
                                     & LEB               & 25.20 hrs    & 155.76 K  &               &            & $\checkmark$\\ 
                                     & MOR               & 8.60 hrs     & 58.38 K   &               &            & $\checkmark$\\ 
                                     & MSA               & 612.28 hrs   & 3.80 M    &               &            & $\checkmark$\\ 
                                     & PAL               & 6.17 hrs     & 45.35 K   &               &            & $\checkmark$\\ 
                                     & SAU               & 452.24 hrs   & 301.92 K  &               &            & $\checkmark$\\ 
                                     & SYR               & 211.33 hrs   & 1.06 M    &               &            & $\checkmark$\\ 
                                     & TUN               & 12.17 hrs    & 34.34 K   &               &            & $\checkmark$\\ 
                                     & UAE               & 9.87 hrs     & 6.42 K    &               &            & $\checkmark$\\ \hline
        MGB2                         & Mostly MSA        & 1000 hrs     & 7.31 M    & $\checkmark$    & $\checkmark$ & $\checkmark$ \\ \hline
        QASR                         & Mostly MSA        & 2000 hrs     & 13.33 M   &               & $\checkmark$ &  \\ \hline
        MGB3$\texttt{[ASR]}$         & EGY               & 2.83 hrs     & 18.93 K   &               & $\checkmark$ &  \\
        MGB3$\texttt{[ADI]}$         & EGY               & 11.15 hrs    & ---       &               & $\checkmark$ & $\checkmark$ \\
                                     & GLF               & 8.92 hrs     & ---       &               & $\checkmark$ & $\checkmark$ \\
                                     & LAV               & 9.27 hrs     & ---       &               & $\checkmark$ & $\checkmark$ \\
                                     & MSA               & 9.39 hrs     & ---       &               & $\checkmark$ & $\checkmark$ \\
                                     & NOR               & 9.49 hrs     & ---       &               & $\checkmark$ & $\checkmark$ \\ \hline
        MGB5$\texttt{[ASR]}$         & MOR               & 115.7hrs     & ---       &               & $\checkmark$ &  \\
        MGB5$\texttt{[ADI]}$         & ALG               & 115.7hrs     & ---       &               & $\checkmark$ & $\checkmark$ \\
                                     & EGY               & 451.1 hrs    & ---       &               & $\checkmark$ & $\checkmark$ \\
                                     & IRA               & 815.8 hrs    & ---       &               & $\checkmark$ & $\checkmark$ \\
                                     & JOR               & 25.9 hrs     & ---       &               & $\checkmark$ & $\checkmark$ \\
                                     & KSA               & 186.1 hrs    & ---       &               & $\checkmark$ & $\checkmark$  \\
                                     & KUW               & 108.2 hrs    & ---       &               & $\checkmark$ & $\checkmark$ \\
                                     & LEB               & 116.8 hrs    & ---       &               & $\checkmark$ & $\checkmark$  \\
                                     & LIB               & 127.4 hrs    & ---       &               & $\checkmark$ & $\checkmark$ \\
                                     & MAU               & 456.4 hrs    & ---       &               & $\checkmark$ & $\checkmark$ \\
                                     & MOR               & 57.8 hrs     & ---       &               & $\checkmark$ & $\checkmark$ \\
                                     & OMA               & 58.5 hrs     & ---       &               & $\checkmark$ & $\checkmark$  \\
                                     & PAL               & 121.4 hrs    & ---       &               & $\checkmark$ & $\checkmark$ \\
                                     & QAT               & 62.3 hrs     & ---       &               & $\checkmark$ & $\checkmark$  \\
                                     & SUD               & 47.7 hrs     & ---       &               & $\checkmark$ & $\checkmark$ \\
                                     & SYR               & 119.5 hrs    & ---       &               & $\checkmark$ & $\checkmark$ \\
                                     & UAE               & 108.4 hrs    & ---       &               & $\checkmark$ & $\checkmark$ \\
                                     & YEM               & 53.4 hrs     & ---       &               & $\checkmark$ & $\checkmark$  \\ \hline
        Mixat$^{\texttt{[cs]}}$                        & UAE               & 9.97 hrs     & 57.82 K   &               &            & $\checkmark$\\ \hline
        ParallelCorp                 & EGY               & 32 hrs       & 48.56 K   &               &            & $\checkmark$\\  
                                     & GLF               & 32 hrs       & 27.26 K   &               &            & $\checkmark$\\ 
                                     & LEV               & 32 hrs       & 18.43 K   &               &            & $\checkmark$\\ 
                                     & MSA               & 32 hrs       & 30.66 K   &               &            & $\checkmark$\\ \hline
        \multirow{3}{*}{MADAR}       & MOR ALG TUN       & \multirow{3}{*}{---}     & \multirow{3}{*}{532.37K}   &               & \multirow{3}{*}{$\checkmark$} & \multirow{3}{*}{$\checkmark$}\\ 
                                     & LIB EGY LEV \\
                                     & IRA GLF YEM \\ \hline
        \multirow{6}{*}{NADI}        & ALG BAH EGY       & \multirow{6}{*}{---}     & \multirow{6}{*}{702.67K}   &               & \multirow{6}{*}{$\checkmark$} & \multirow{6}{*}{$\checkmark$} \\ 
                                     & IRA JOR KUW \\
                                     & LEB LIB MOR \\
                                     & OMN PAL QAT \\
                                     & SAU SUD SYR \\
                                     & TUN UAE YEM \\ \hline
        SADA                         & SAU               & 417.63 hrs   & 3.26M     &               &           & $\checkmark$\\ \hline
        TARIC-SLU                    & TUN               & 6.81 hrs     & 53.48K    &               &           & $\checkmark$ \\ \hline
        TunSwitch$^{\texttt{[cs]}}$  & TUN               & 10.89 hrs    & 70.86K    &               &           & $\checkmark$ \\ \hline
        VoxBlink                     & Dialect Mix       & 19.92 hrs    &   ---  &               &           & $\checkmark$  \\
        \bottomrule
     \end{tabular}}
     \caption{Summary of Dataset Statistics for \textbf{Pre-Training}: Hours of Audio, Word Counts, and Associated Dialects. *$^{\texttt{[cs]}}$ refers to Code Switching datasets. *$^{\texttt{[txt]}}$ refers to textual datasets.}
     \label{tab:dataset_summary_combined_pretraining}
 \end{table}

\section{Inference examples}

\begin{figure}[h]
    \centering
    \includegraphics[width=1\linewidth]{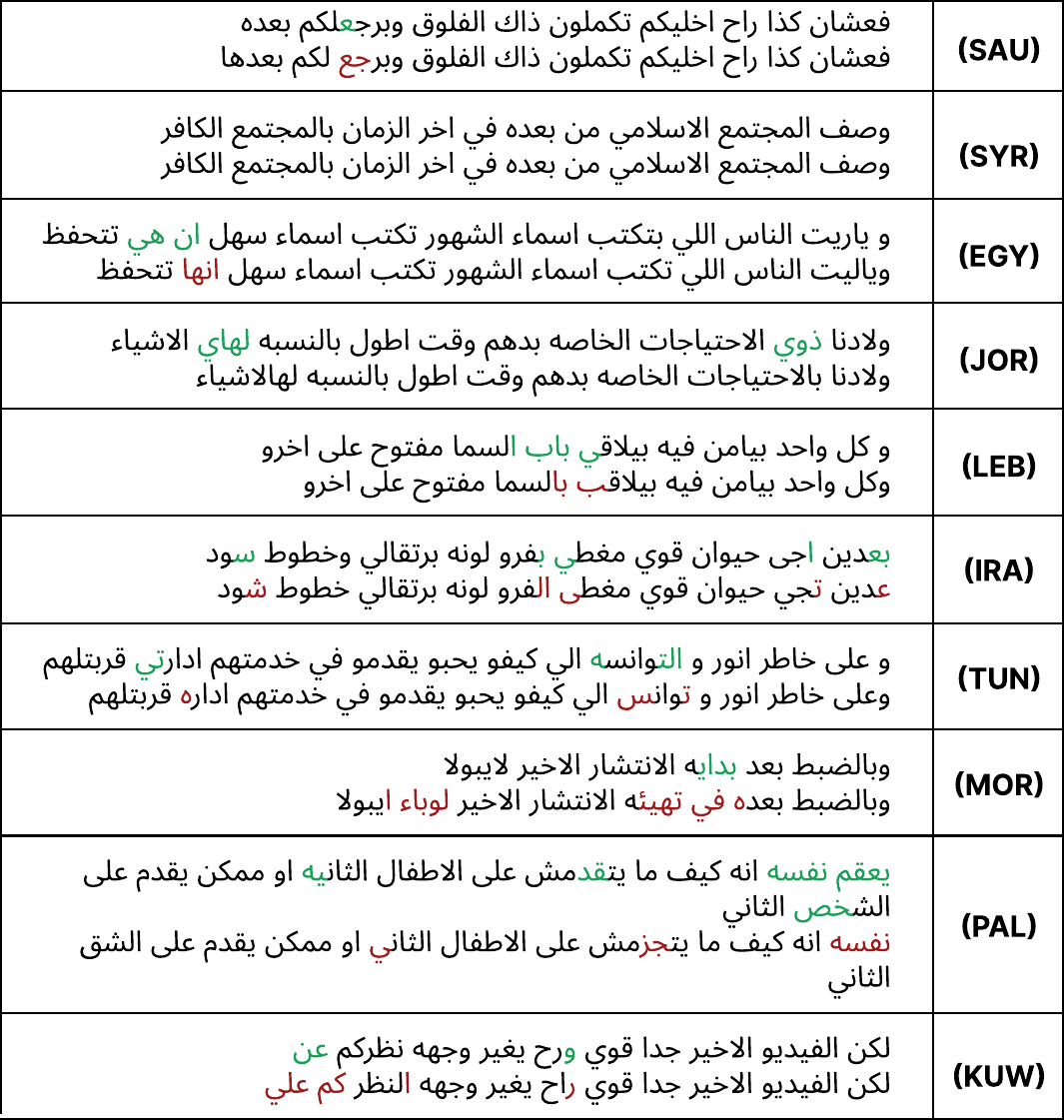}
    \caption{Examples of dialectal recognition after targeted fine-tuning, following MGB2 adaptation.}
    \label{fig:enter-label}
\end{figure}

Figure~\ref{fig:enter-label} lists examples of ASR outputs using the dialect-specific fine-tuned models. Note that the `errors' in SAU, EGY, and JOR examples are in fact alternative spellings.

\begin{figure*}
    \centering
    \includegraphics[width=1.7\columnwidth]{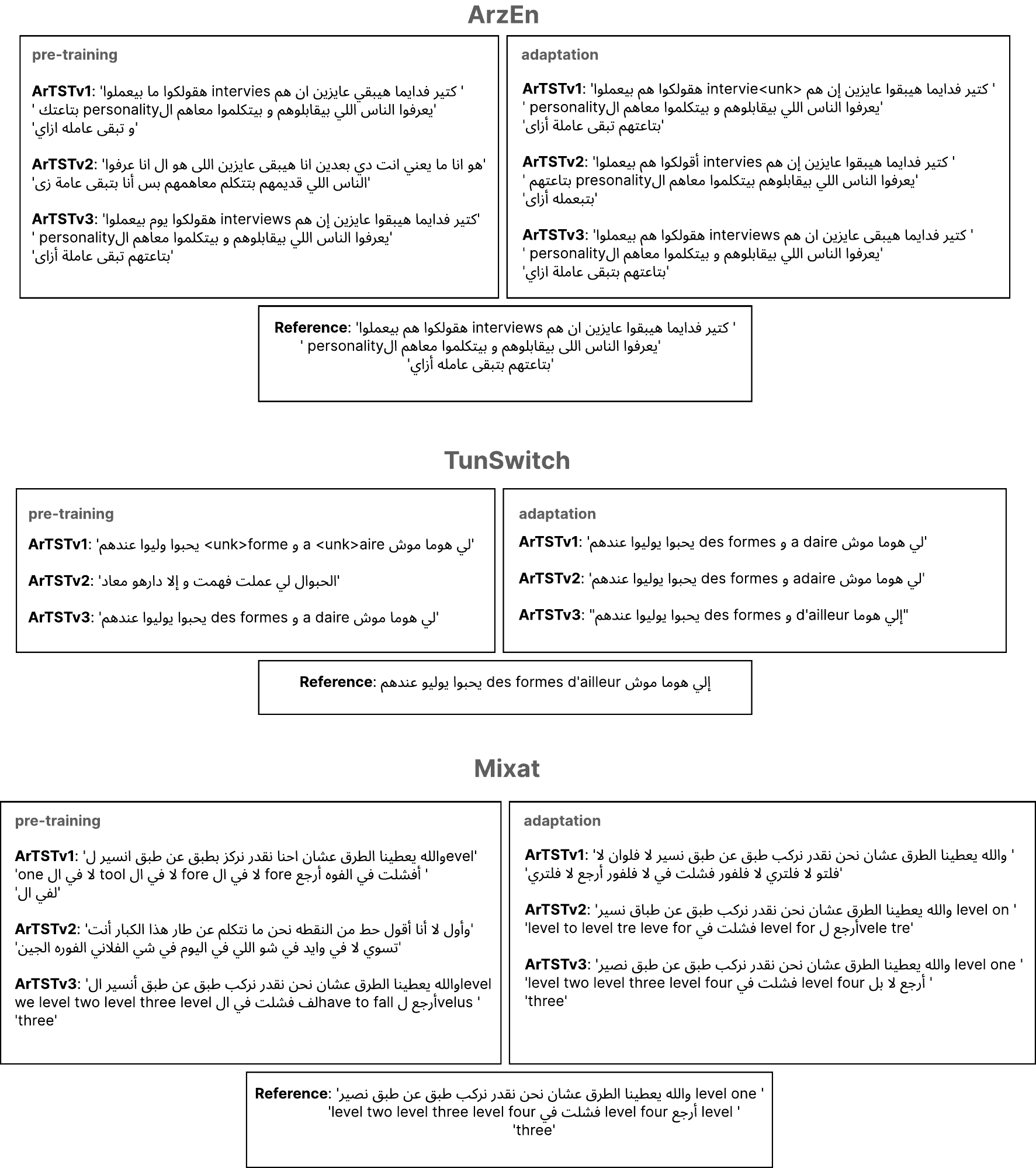}
    \caption{Examples of code-switching samples from each of the code-switching datasest using (a) pre-training checkpoints fine-tuned on CS data directly, and (2) adapted models after fine-tuning on general data: v1 (MGB2), v2 (Dialectal data from Table~\ref{tab:joint_dialect_training_stata}), and v3 (dialectal data from Table~\ref{tab:joint_dialect_training_stata} + Multilingual and CS data from Table~\ref{tab:joint_multilingual_data}).}
    \label{fig:code-switching-examples}
\end{figure*}

\begin{table*}
\centering
\resizebox{2\columnwidth}{!}{
    \begin{tabular}{lc|ccccccccccc}
        \toprule
         & \multirow{2}{*}{\textbf{Dialect}} & \multicolumn{11}{c}{\textbf{Fine-Tuning Set}} \\
         & & \textbf{SAU(\%)} & \textbf{SYR(\%)} & \textbf{EGY(\%)} & \textbf{JOR(\%)} & \textbf{LEB(\%)} & \textbf{IRA(\%)} & \textbf{TUN(\%)} & \textbf{MOR(\%)} & \textbf{PAL(\%)} & \textbf{KUW(\%)} & \textbf{UAE(\%)} \\
        \hline
        \multirow{10}{*}{\textbf{test set}}
        & \textbf{SAU} & \diagcell{\textbf{27.33}} & 53.88 & 63.24 & 43.95 & 56.64 & 46.10 & 49.36 & 50.02 & 45.48 & 46.85 & 46.14 \\
        & \textbf{SYR} & 19.54 & \diagcell{17.42} & 23.91 & 16.88 & 25.71 & \textbf{16.85} & 24.36 & 17.58 & 16.86 & 17.38 & 17.53 \\
        & \textbf{EGY} & 38.07 & 53.43 & \diagcell{\textbf{36.43}} & 40.08 & 51.61 & 41.37 & 45.58 & 43.88 & 40.41 & 41.88 & 40.38 \\
        & \textbf{JOR} & 22.88 & 30.37 & 34.14 & \diagcell{\textbf{21.08}} & 28.11 & 28.68 & 32.96 & 30.50 & 25.25 & 27.88 & 28.05 \\
        & \textbf{LEB} & 41.03 & 42.53 & 53.23 & 39.07 & \diagcell{\textbf{28.05}} & 41.23 & 48.53 & 42.34 & 41.55 & 42.14 & 40.97 \\
        & \textbf{IRA} & \textbf{35.18} & 49.10 & 56.91 & 40.47 & 46.72 & \diagcell{36.10} & 47.84 & 42.55 & 40.76 & 41.00 & 42.16 \\
        & \textbf{TUN} & 44.44 & 57.78 & 51.85 & 46.67 & 45.19 & 45.19 & \diagcell{\textbf{26.67}} & 44.44 & 47.41 & 46.67 & 47.41 \\
        & \textbf{MOR} & 59.04 & 71.69 & 74.70 & 54.82 & 69.88 & 57.23 & 74.70 & \diagcell{56.63} & \textbf{56.02} & 59.64 & 57.83 \\
        & \textbf{PAL} & \textbf{48.53} & 66.18 & 62.65 & 52.35 & 60.59 & 58.53 & 60.00 & 63.24 & \diagcell{53.53} & 58.53 & 60.59 \\
        & \textbf{KUW} & \textbf{26.59} & 59.54 & 78.03 & 43.93 & 67.63 & 46.82 & 53.18 & 51.45 & 44.51 & \diagcell{46.24} & 50.87 \\
        \hline
        \multirow{3}{*}{\textbf{held-out}} 
        & \textbf{ALG} & \textbf{50.21} & 60.09 & 73.61 & 52.58 & 62.02 & 57.51 & 57.51 & 59.01 & 52.79 & 58.37 & 57.30 \\
        & \textbf{SUD} & \textbf{40.89} & 64.97 & 64.32 & 53.15 & 64.86 & 54.12 & 56.51 & 58.79 & 52.39 & 56.18 & 55.53 \\
        & \textbf{YEM} & 38.40 & 42.16 & 38.49 & 34.68 & 39.69 & 34.92 & 30.68 & 35.73 & \textbf{33.06} & 34.68 & 34.11 \\
        \bottomrule
    \end{tabular}
}
\caption{WER(\%) for various models on unseen dialects. All models are adapted from v2$\rightarrow$MGB2.}
\label{tab:zero_shot_cx_dialects_performance}
\end{table*}



\noindent Figure~\ref{fig:code-switching-examples} shows example outputs from each model on the code-switching datasets: ArzEn (Egyptian-English), TunSwitch (Tunisian-French), and Mixat (Emirati-English). 




\section{Cross-Dialectal Performance}
Table \ref{tab:zero_shot_cx_dialects_performance} shows the cross-dialectal performance, where models trained on a single target dialect are tested on other dialects, including the three held-out sets: ALG, SUD, and YEM.
In most cases, the best performance is achieved by the model trained on the same target dialect (the diagonal in Table \ref{tab:zero_shot_cx_dialects_performance}). However, for low-resource dialects, like KUW and PAL, the model trained on SAU achieved the lowest WER. This is likely a result of the large size of the SAU train set and the wide geographical area and dialectal variants it covers. Curiously, all models perform well on the SYR test set; upon close inspection, we found that the set consists mostly of MSA utterances, which explains the result since all models are adapted from MSA.


\end{document}